# Formulation of Deep Reinforcement Learning Architecture Toward Autonomous Driving for On-Ramp Merge

Pin Wang, Ching-Yao Chan
California PATH, University of California, Berkeley
Berkeley, US
{pin_wang, cychan}@berkeley.edu

*Abstract*— Multiple automakers have in development or in production automated driving systems (ADS) that offer freeway-pilot functions. This type of ADS is typically limited to restricted-access freeways only, that is, the transition from manual to automated modes takes place only after the ramp merging process is completed manually. One major challenge to extend the automation to ramp merging is that the automated vehicle needs to incorporate and optimize long-term objectives (e.g. successful and smooth merge) when near-term actions must be safely executed. Moreover, the merging process involves interactions with other vehicles whose behaviors are sometimes hard to predict but may influence the merging vehicle's optimal actions. To tackle such a complicated control problem, we propose to apply Deep Reinforcement Learning (DRL) techniques for finding an optimal driving policy by maximizing the long-term reward in an interactive environment. Specifically, we apply a Long Short-Term Memory (LSTM) architecture to model the interactive environment, from which an internal state containing historical driving information is conveyed to a Deep Q-Network (DQN). The DQN is used to approximate the Q-function, which takes the internal state as input and generates Q-values as output for action selection. With this DRL architecture, the historical impact of interactive environment on the long-term reward can be captured and taken into account for deciding the optimal control policy. The proposed architecture has the potential to be extended and applied to other autonomous driving scenarios such as driving through a complex intersection or changing lanes under varying traffic flow conditions.

## I. INTRODUCTION

Highly or fully automated systems, such as Tesla Autopilot [1] and google self-driving car [2], are not widely available yet from major automakers. Some automakers are likely to offer partially or highly automated features at Automation Level 2, 3 or 4 in the very near future, such as those provided by Volvo [3] in the Swedish "Drive Me" experiment. Despite the advancement of high automation levels being proposed and demonstrated, the implementation of autonomous driving for highway on-ramp merge still presents considerable challenges. First, the ADS needs to decide on immediate actions with the consideration of delayed impacts on the future vehicle states. For example, the actions such as accelerating, decelerating, or steering that the ego vehicle takes at the current time will affect the success or failure of the merge mission. This process can be handled at relative ease in most cases by experienced human drivers but the algorithms for automatic execution of the merge maneuver in a consistently smooth, safe, and reliable manner can become complex. Second, the ego vehicle's merge maneuver depends not only on its own states and actions but also on its interaction with surrounding vehicles, which may be cooperative or adversarial. A typical scenario is that a vehicle on the mainline arriving at the merge point from behind may operate cooperatively (e.g. decelerate or change lane) to let the merging vehicle merge or it may act adversarially (e.g. speed up) to deter the merging vehicle from entering into the mainline traffic. Such interaction is not trivial, and serious risks may emerge if the ego vehicle fails to respond to the potentially complicated interaction.

In this paper, we propose a machine learning framework, Deep Reinforcement Learning, to achieve a robust and reliable merging policy. With this approach, it is of vital importance to learn the interactive environment and to optimize long term cumulative rewards. In our study, we formulate an architectural framework based on Deep Reinforcement Learning techniques for the on-ramp merge problem to tackle the issues of (1) attaining long-term effects, (2) mitigating unexpected and adversarial impacts by other agents, and (3) dealing with a system of continuous states/actions.

A literature view of related works is described in the next section, followed by the proposed architecture and the methodology. Then, the implementation procedure of the ramp-merge problem is presented. Finally, concluding remarks and discussions are given in the closing section.

## II. LITERATURE REVIEW

Several modeling methods have been previously suggested to solve the autonomous on-ramp merging problem by assuming some specific rules. Davis [4] presented a cooperative merging strategy, in which vehicles on the mainline always slow down to create enough gap space to let the on-ramp vehicle merge into. Marinescu et al. [5] proposed a slot-based merging algorithm, which defined a slot's occupancy status (e.g. free and occupied) based on the mainline vehicles' speed, position, and behavior of acceleration or deceleration. Chen et al. [6] used driving rules and a gap acceptance theory to model the decision-making process of the urban expressway on-ramp merge problem. These rule-based models are conceptually comprehensible but are pragmatically vulnerable due to their inability to adapt to unforeseen situations in the real world. In other words, merge maneuvers can only be executed under predefined rules, and for cases that are outside the domains of these rules, the models may fail and lead to hazardous situations, e.g. crashes or severe disturbance to traffic flows.

Machine learning models, in contrast, have the potential to be superior in dealing with complex situations without resorting to detailed hard-coded rules or pre-determined models. Particularly, reinforcement learning, different from standard supervised learning techniques, which need ground truth input/output pairs, can efficiently learn optimal actions by itself through trials and errors [7]. A reinforcement learning agent observes its environment, and interacts with it by taking actions that are balanced between exploration of uncharted territory and exploitation of current knowledge. Then it receives immediate rewards and the system moves to a new state. By maximizing the cumulative reward, the agent finds an optimal policy to achieve its goal.

Reinforcement learning has been extensively applied to the field of robotics and recently been applied to vehicle and traffic control problems. Fares et al. [8] designed a Reinforcement Learning based Density Control Agent (RLCA) to control the number of vehicles entering the mainline from the ramp merging area. Yang et al. [9] developed a ramp-metering control algorithm based on reinforcement learning to increase the capacity at weaving sections. These applications use the basic Q-learning in which the state space and action space are discrete and the problem is considered as a Markov Decision Process (MDP).

Ramp merging is much more complex. The driving environment includes not only the merging vehicle's state but also the dynamic states of other agents, which are not necessarily predictable in the view of the merging agent. Therefore, the on-ramp merge case is an intrinsically non-Markovian problem. Moreover, the vehicle's state space and action space are continuous, which makes it impractical to use tabular settings as in basic Q-learning. Instead, Q-function approximation is a good way to deal with non-MDP or Partially Observed Markov Decision Process (POMDP) in a continuous state/action space. For example, Google DeepMind [10] has successfully applied a Deep Q-network (a convolutional neural network, CNN) to play Atari games with only screen images and game scores as inputs. Some other studies applied a similar reinforcement learning framework on browser-based car simulators to implement autonomous vehicle control under some specific scenarios. For instance, Sallab et al. [11] used end-to-end deep reinforcement learning for lane-keeping assist on an open-source simulator for Racing called (TORCS). Yu et al. [12] investigated the use of deep reinforcement learning for training an agent to control a simulated car running on the track in JavaScript Racer. In these applications, the inputs of game screens only represent simplified real-world driving rules so that the policy learned can only be applied on virtual video games. Mobileye [13] used another approach that employed two supervised learning models to describe an interactive environment and a recurrent neural network based on these two models to learn an optimal policy. Shalev-Shwartz et al. [14] decompose the driving strategy into a learnable part which estimate the comfort of driving, and a non-learnable part which is hard constrains on the safety of driving.

The aforementioned research and associated shortcomings show that basic Q-learning is not appropriate for handling problems with non-MDP properties and continuous states/actions, and that Deep Q-network with CNN structure is limited to image inputs and not ready to be implemented on real-world driving scenarios. Besides, the historical driving information has not been extensively incorporated in these studies. It is important to develop a highly representative model to describe the real-world environment and design an appropriate Q-function approximator to employ long dependencies of the history for learning a robust and reliable driving policy.

III. METHODOLOGY

In our study, we use Deep Reinforcement Learning to incorporate the influence of the historical information of the driving environment on the merging policy optimization. Specifically, we model the environment with a Long Short-Term Memory (LSTM) architecture to learn the internal relations between the ego vehicle and other surrounding vehicles based a relatively long duration of the past time. An internal state representation from LSTM at each time step is then fed into a Deep Q-network for action selection. After that, the Q-network is immediately updated by an experience replay and a second target Q-network to avoid local optima and divergence problems. In this way, an interactive merging policy can be learned. The overview of the architecture is shown in Fig. 1. We will first introduce the architecture of LSTM and then the architecture of Deep Q-learning.

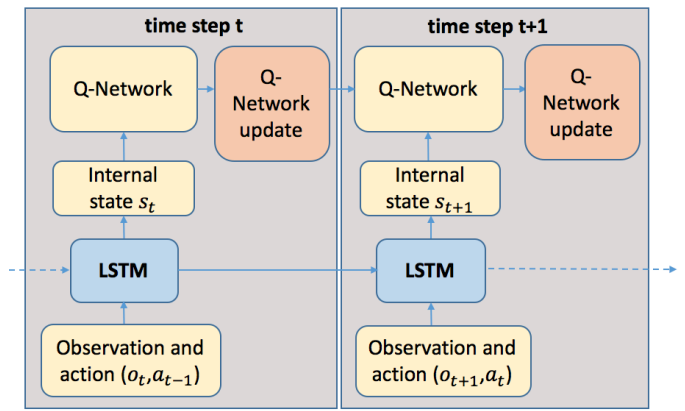

Figure. 1 Architecture overview of Deep Reinforcement Learning

*A. LSTM*

LSTM is a special recurrent neural network designed to handle long-term dependency problems [15]. LSTM has the ability to remember values for both long and short durations. As mentioned earlier, the driving environment of the on-ramp merge scenario involves interactions with the surrounding vehicles, which are not necessarily predictable in the view of the ego vehicle. With the use of LSTM, the historical driving information can be incorporated into an internal state that is an adequate representation of the interactive environment. In other words, the internal state given by an LSTM cell gives a compact and fixed-sized representation of the history that can be fed into the Q-network.

In our study, we train the LSTM model by supervised learning. The architecture of the LSTM model is shown in Fig. 2. Each LSTM unit includes two modules, an observation module and a state module. The observation module is used to estimate the next observation ($\hat{o}_i$) based on the inputs of the internal state ($s_{i-1}$) and action ($a_{i-1}$) from the last time step. The state module is used to map the current observation ($o_i$) into a more informative internal state ($s_i$) by employing historical environment information retained from previous steps. The representative internal state is then input to the next LSTM unit as well as to Q-network for action decision, which is described in the following section.

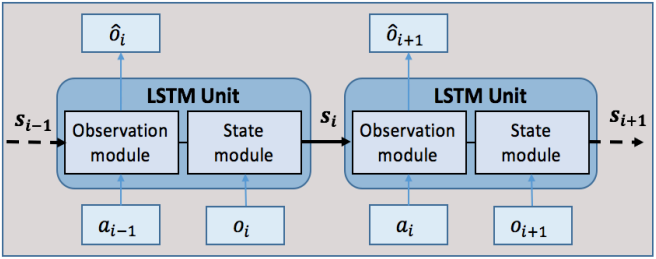

Figure. 2 LSTM architecture

*B. Deep Q-learning*

Deep Q-learning is a model-free approach to deal with reinforcement learning problems. Deep Q-learning uses neural networks, parameterized by θ, to approximate the Q-function. Q-values, denoted as $Q(s,a;\theta)$, can be used to get the best action for a given state. The architecture of Deep Q-learning in our study is depicted in Fig. 3.

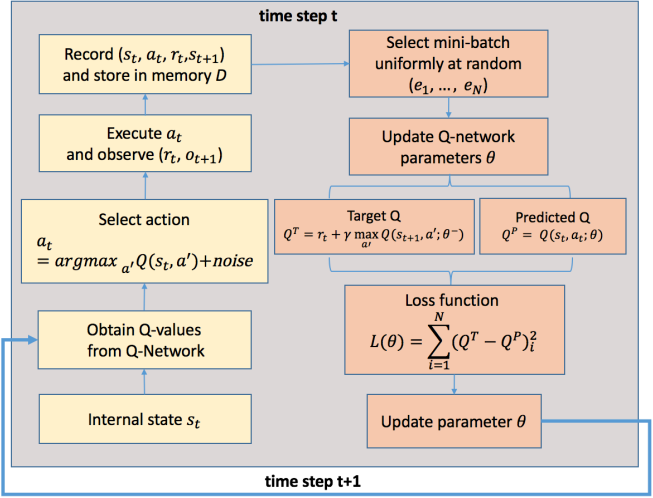

Figure. 3 Deep Q-learning Architecture

In our approach, the Q-learning process consists of two parts at each time step. One is the Q-value approximation for action selection (left part in Fig. 3) in which the internal state $s_t$, obtained from LSTM based on $o_t$, is used as the input to the Q-network to get the chosen action $a_t$. The other part is the Q-network update (right part in Fig.3) where the loss between predicted Q-values and target Q-values is used to update Q-network parameters $\theta$ and $\theta^-$.

To guarantee there is always an optimal action in a given state and that the learning process can converge fast, in our work, we design the Q-function approximator as a quadratic function, $Q(s,a) = A(s) * (B(s) - a) ** 2 + C(s)$, where $A$, $B$, and $C$ are designed with neural networks.

The vehicle action is composed of the longitudinal control (acceleration) and the lateral control (steering). Note that the action cannot be arbitrarily large or small values due to the vehicle physical mechanics, therefore, we restrain the acceleration and steering angle in certain ranges, and within the range the acceleration and steering can be any real value. When the best action is obtained with the highest Q-value, a random noise is added to it and the new value is the chosen action, which is similar with the concept of $\varepsilon$-greedy used in the action exploration process.

$$a_t = \arg\max_{a'} Q(s_t, a') + noise \qquad (1)$$

The reward function measures the safeness, smoothness, and timeliness of the merging maneuver, and is formulated as a function of the merging vehicle's acceleration, steering angle, speed, and the distance to its surrounding vehicles, shown in equation (2).

$$R(s_t, a_t) = r1 * accel + r2 * steering + r3 * distance + r4 * speed \qquad (2)$$

where $r1$ and $r2$ are the smoothness factors; $r3$ is the safeness factor; and $r4$ is the timeliness factor. Large acceleration/deceleration, small distance to surrounding vehicles, and low speed under free flow condition are penalized by large negative rewards.

The loss function is defined by the mean square error between predicted Q-values $Q_t^P$ and target Q-values $Q_t^T$, equation (3). $Q_t^T$ is calculated by the immediate reward $r_t$ and the maximum Q-value of the next internal state $s_{t+1}$.

$$L(\theta) = \sum (Q_t^T - Q_t^P)^2 \qquad (3)$$

where $Q_t^P = Q(s_t, a_t|\theta)$, $Q_t^T = r_t + \gamma \max_{a'_{t+1}} Q(s_{t+1}, a'_{t+1}|\theta^-)$. $\gamma$ is a discounted factor, $\gamma \in [0,1]$ The Q-network parameters are updated by the backpropagation on $L(\theta)$:

$$\theta := \theta + \nabla_\theta L(\theta) \qquad (4)$$

It is worth mentioning that there may be stability issues with such a Q-learning approach due to the fact that the data we used in the Q-network training is sequential and that slight changes in Q-values can lead to large oscillations in control policies. With these considerations, we apply an experience replay and use a different set of Q-function parameters $\theta^-$ for $Q_t^T$, to break data correlation and to avoid sticking in local optima. At each time step $t$, a state transition element can be saved as $e_t = (s_t, a_t, r_t, s_{t+1})$ after the action execution, and is stored in a replay memory $D$. A mini-batch is sampled uniformly at random as $\{e_1, ..., e_N\}$, which are used for the parameter updates as in supervised learning. The parameters used for the estimation of target Q-values are fixed for a number of iterations (e.g. $C$ iterations where $C=500$), and updated periodically with the Q-network parameters at that step, as $\theta^- \leftarrow \theta$.

IV. IMPLEMENTATION ON ON-RAMP MERGE CASE

The implementation of the proposed Deep Reinforcement Learning architecture for the highway on-ramp merge problem is described below.

A typical scenario of the on-ramp merging involves three vehicles. The merging vehicle on the ramp, denoted as $V_m$, executes the merge maneuver when it observes an appropriate gap formed by two vehicles on the mainline, denoted as the gap lag vehicle $V_l$ and the gap front vehicle $V_f$, as shown in Fig. 4(a) (a simulated scenario of a section of US Interstate Highway I-80). The training is based on such simulated scenarios of real-world merging areas.

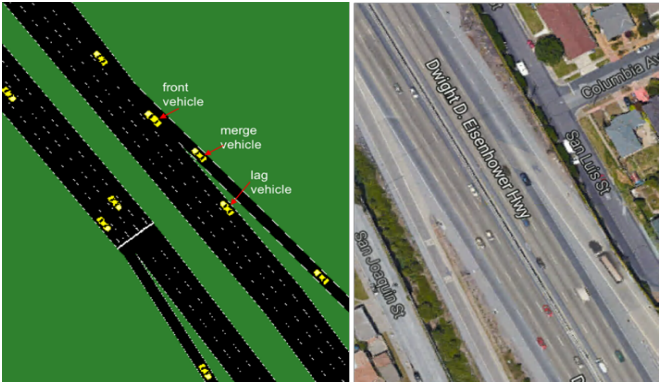

Figure. 4 Ramp merge ((a) simulated scenario and (b) real-world location)

In the current study, we only consider the interaction among the three vehicles (the merging vehicle, the gap lag vehicle and the gap front vehicle). For the merging vehicle, we use 5 variables to describe its driving state that are speed ($v^m$), position ($p^m$), heading angle ($\varphi^m$), and distances to the right ($r^m$) and left ($l^m$) lane makings of the current lane. For the other two vehicles, we assume we can only observe their speeds ($v^f$ and $v^l$) and positions ($p^f$ and $p^l$). In total, there are 9 variables in the state representation. The observation at time step $t$ is denoted as $o_t = (v_t^m, p_t^m, \varphi_t^m, l^m, r^m, v_t^f, p_t^f, v_t^l, p_t^l)$.

The action is the acceleration and the steering angle taken by the merge vehicle, and is denoted as $a_t = (a_t^m, \sigma_t^m)$. The immediate reward is measured by the $o_t$ and $a_t$. The threshold for action variables and for negative rewards can be fine-tuned for specific design criteria.

To guarantee the representativeness of the driving environment, the data to train the LSTM model are obtained from the real-world driving data. Cameras, placed on a high pole, are used to record the on-ramp merging scenarios in a bird view. Video analysis (e.g. scene understanding, object detection, motion estimation) is performed to obtain the aforementioned observation and action variables. In the video analysis, the inputs are video images and the output are vehicle dynamics. The interactive driving behaviors of the merge vehicle and the other two vehicles (the lag vehicle and the front vehicle) are learned as internal state through the LSTM units. In the Deep Q-learning process, a couple of neural networks are used to calculate Q values based on the internal state at each time step. The parameter update of the Q-network is conducted through gradient descent. The pseudo codes are shown below.

**Algorithm 1 Deep Q-learning with experience replay**

1: Initialize experience replay memory $D$
2: Initialize Q-Network parameters $\theta$
3: for episode $p = 1: M$ do
4:    Initialize environment state $o_1$
5:     for time $t = 1: T$ do
6:       get the initial state $s_t$ from LSTM
7:       input $s_t$ and $a_t$ to Q-function approximator
8:       get the best action $a_t^*$
9:       get the chosen action $a_t = a_t^* + noise$
10:      execute $a_t$ and obtain immediate reward $r_t$ and next observation $o_{t+1}$
11:      store transition element $(s_t, a_t, r_t, s_{t+1})$ into $D$
12:      sample mini-batch $(e_1, \ldots, e_N)$ uniformly at random from $D$ ($e_i = (s_i, a_i, r_i, s_{i+1}), \forall i \in [1, N]$)
13:      for $i = 1: N$ do
14:        set target Q as $Q_i^T$
$$= \begin{cases} r_i, & \text{if } s_{i+1} \text{ is terminal state} \\ r_i + \gamma * \max_{a'}(Q(s_{i+1}, a'; \theta^-)), & \text{if } s_{i+1} \text{ is not terminal state} \end{cases}$$
15:       calculate $L(\theta) = L(\theta) + (Q_i^T - Q(s_i, a_i; \theta))^2$
16:      perform gradient decent on $L(\theta)$
17:      set $\theta^- = \theta$ after $C$ iterations

This paper describes our preliminary work on autonomous ramp merging. The verification and validation of the implemented methodologies is our next step work.

## V. CONCLUSION AND DISCUSSION

In this work, we propose a Deep Reinforcement Learning architecture for learning an on-ramp merge driving policy. The driving environment is based on a LSTM architecture to incorporate the influence of historical and interactive driving behaviors on the action selection. In the Deep Q-learning process, the internal state from LSTM is taken as the input to the Q-function approximator, which is used for the action selection based on more past information. The Q-network parameters are updated with an experience replay and a second target Q-network is used to relieve the problems of local optima and instability. Model training and fine-tuning, refinement of proposed methodologies, and performance evaluation of deep learning approaches remain topics of our further studies.

We also envision that there is another research topic, which we call "supervisory control in highly automated driving systems", that is worth pursuing. A sensible operational concept of ADS is a system, in which manual and automated modes may cooperatively co-exist, instead of forced switching between the two modes. The "supervisory control" term per Sheridan [16], suggests that human-machine systems can exist in a spectrum of automation, and shift across the spectrum of control levels in real time to suit the situation at hand. In the use case of ramp merge, the driver may selectively provide control inputs while the automated driving system seeks to maximize the reward and achieve a successful and effective merge maneuver. To accommodate a balanced and common reward in the machine learning approach, a cooperative inverse reinforcement learning concept (CIRL) [17] may be considered and adopted into the framework.

ACKNOWLEDGMENT

The authors thank the support from Berkeley Deep Drive program and the helpful discussion with Dr. Yi-Ta Chuang.